\documentclass{article}

\usepackage[logo]{ace_ego}        

\usepackage{amsfonts}
\usepackage{dsfont}
\usepackage{amsmath}
\usepackage{graphicx}
\usepackage{algorithm}
\usepackage{algpseudocode}
\usepackage{booktabs}
\usepackage{enumitem}
\usepackage{microtype}
\usepackage{afterpage}
\usepackage{xcolor}
\usepackage{marvosym}
\title{Hierarchical Advantage Weighting for Online RL Fine-Tuning of VLAs from Sparse Episode Outcomes}

\date{Jun 2026}
\metadata[Website]{\url{https://acerobotics-vla.github.io/HABC-Website}}
\metadata[Code]{\url{https://github.com/ACERobotics-VLA/HABC}}

\author{%
  \sffamily\bfseries\small
  Tongyan Fang$^{1,2}$ \And
  Siyuan Huang$^{1\dagger}$ \And
  Naiyu Fang$^{1,3}$ \And
  Ganlong Zhao$^{1,3}$ \And
  Zhongjin Luo$^{1,3}$ \And
  Jianbo Liu$^{1}$ \AND
  Xiaogang Wang$^{1}$ \And
  Ying Dong$^{2{\,\scalebox{0.75}{\normalfont\Letter}}}$ \And
  Hongsheng Li$^{1,3{\,\scalebox{0.75}{\normalfont\Letter}}}$
  \\[6pt]
  \mdseries\normalsize
  \begin{tabular}[t]{@{}l@{}}
    $^{1}$ACE Robotics \quad
    $^{2}$Shenzhen International Graduate School, Tsinghua University \\
    $^{3}$The Chinese University of Hong Kong\\[3pt]
    {\footnotesize
    $^\dagger$Project leader \quad
    $^{{\scalebox{0.75}{\normalfont\Letter}}}$Corresponding authors}
  \end{tabular}
}

\begin{document}
\maketitle


\begin{abstract}
    When pretrained VLA policies are fine-tuned through online RL, each rollout episode produces only a single binary outcome (success or failure), yet the actor update requires per-transition supervision.
    Existing approaches commonly reduce this sparse outcome to a single scalar reward or advantage signal, which conflates distinct forms of transition-level feedback and provides limited guidance once basic task success becomes achievable.
    First, a single scalar signal conflates the two objectives of \emph{viability} and \emph{efficiency}; once basic success is achieved, the binary label provides no gradient to distinguish efficient completions from slow ones.
    Second, real-world rollouts mix autonomous and intervention segments; naively assigning episode outcomes across these boundaries introduces incorrect credit assignment.
    To address these issues, we propose Hierarchical Advantage-Weighted Behavior Cloning (HABC), which trains separate critic heads for these two objectives on different data subsets and combines their outputs with a state-adaptive balance.
    A state-adaptive gate $g_t$ merges their one-step advantages, prioritizing viability when success is uncertain and shifting to efficiency only when viability is high, and converts the result into per-transition weights on the actor loss.
    Intervention-aware credit assignment further restricts outcome labels to segments executed by the current policy, preventing supervision from leaking across intervention boundaries.
    In real-robot experiments on three contact-rich bimanual tasks, HABC raises success from supervised fine-tuning (SFT) baselines of 36\%, 44\%, and 12\% to 92\%, 88\%, and 38\%.

\end{abstract}

\begin{figure*}
    \includegraphics[width=\textwidth]{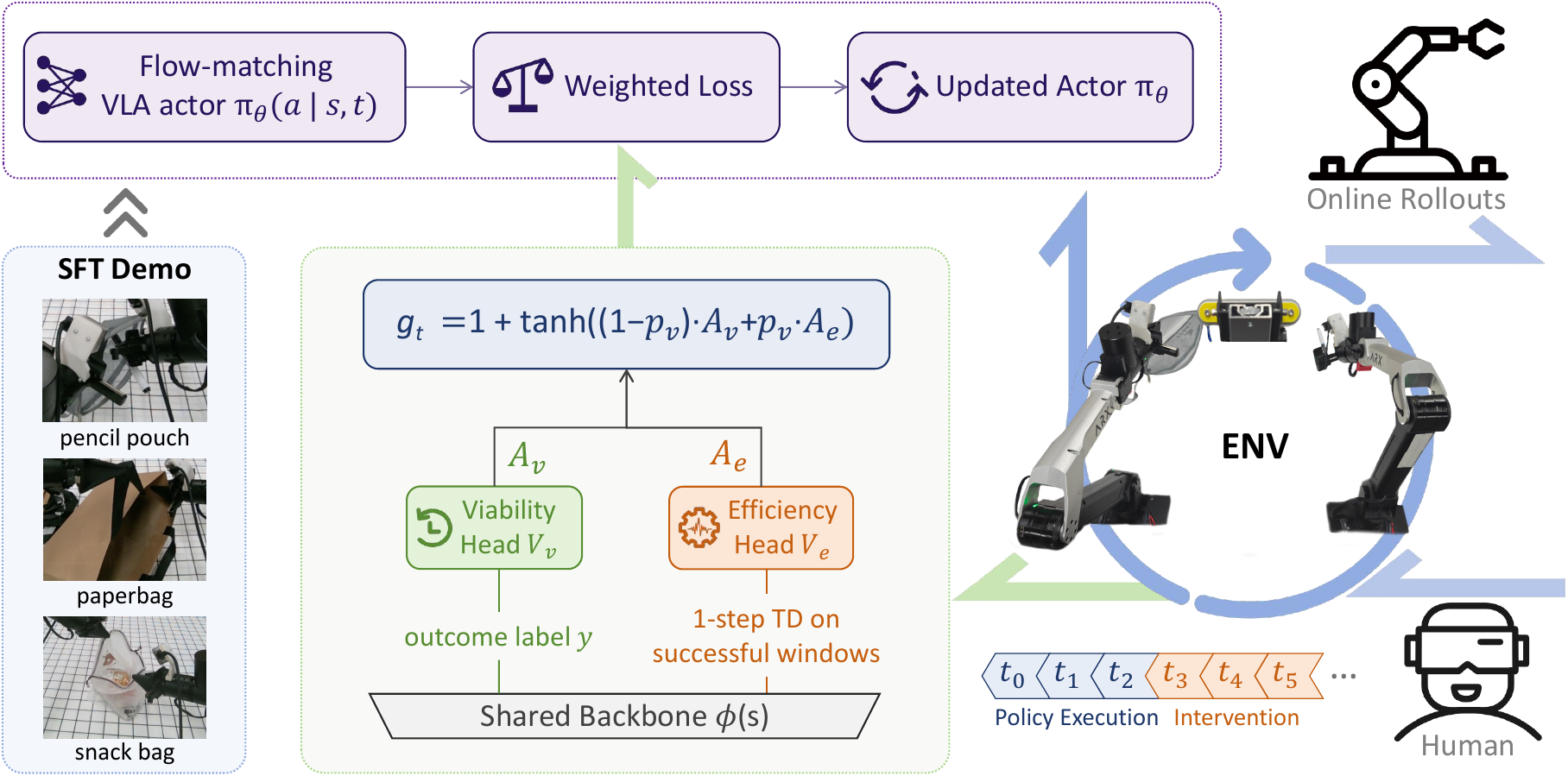}
    \setlength{\abovecaptionskip}{0.5pt} 
    \caption{\textbf{Overview of Hierarchical Advantage-Weighted Behavior Cloning.}
    HABC fine-tunes a VLA actor using SFT demonstrations and online rollouts with policy execution and human interventions.
    Given sparse episode outcomes, HABC converts rollouts into transition-level weights with a dual-head critic.
    The viability head $V_v$ estimates whether a state can still lead to success, while the efficiency head $V_e$ estimates progress toward faster completion.
    Their one-step advantages, $A_v$ and $A_e$, are combined by a state-adaptive gate $g_t$, emphasizing viability when success is uncertain and efficiency once viability is high.
    Intervention-aware credit assignment partitions rollouts by control authority, preventing outcomes from being credited to policy mistakes or human corrections.
    The resulting per-transition weight $w_i$ reweights the flow-matching imitation loss and updates the actor.}
    \label{fig:method_overview}
\end{figure*}


\section{Introduction}
\label{sec:intro}

Pretrained Vision-Language-Action (VLA) policies~\citep{brohan2022rt1,rt2,octo,openvla,pi_0,liu2025rdt} have demonstrated remarkable generalization across diverse manipulation tasks, but demonstrations alone are often insufficient for reliable deployment.
Supervised fine-tuning (SFT) is bounded by demonstration coverage, and covariate shift causes errors to compound at deployment~\citep{ross2011dagger,mandlekar2021robomimic}.
To correct the mistakes a policy actually makes, improve robustness beyond the level of teleoperation, and adapt to new deployment conditions, online RL fine-tuning is necessary where the policy must learn from its own experience~\citep{luo2025hilserl,chen2025conrft,tan2025riptvla,intelligence2025pi}.
In practice, however, each episode produces only a single binary outcome, yet effective actor updates call for per-transition signals.

We observe that this sparse episode label encodes two separable layers of transition-level information: \emph{viability}---whether the current state can still lead to task completion, and \emph{efficiency}---given that success is reachable, whether the current transition is advancing toward completion or wasting time.
Viability can be supervised from all outcome-labeled windows, while efficiency can only be estimated from successful trajectories.
They are also informative at different training stages: viability dominates early when failures are common, while efficiency matters later when success rate is high.
Existing methods such as Recap~\citep{intelligence2025pi} collapse both into a single reward-derived advantage, and critic-filtered behavior cloning hard-selects transitions above a TD-advantage threshold; both approaches lose the structure that would make each signal useful at the right training stage.

Beyond signal decomposition, a second challenge concerns data quality in mixed-control episodes.
When a human intervenes mid-execution, the episode outcome reflects both policy execution and human intervention; naively assigning this outcome to all timesteps can upweight the policy mistakes that triggered the intervention or penalize the human's corrective actions.
Prior work acknowledges that not all online rollout data is equally reliable~\citep{intelligence2025pi}, but does not explicitly handle the control-authority boundary within an episode.
Restricting outcome labels to policy execution segments provides cleaner supervision while still leveraging the human's corrective actions as imitation data.

We address these two challenges with Hierarchical Advantage-Weighted Behavior Cloning (HABC, Fig.~\ref{fig:method_overview}).
A dual-head critic separates the two signals: a viability head trained on all outcome-labeled windows and an efficiency head trained on successful trajectories only, whose one-step advantages are combined through a state-adaptive gate into bounded per-transition weights on the actor loss.
Intervention-aware credit assignment restricts outcome labels to policy execution segments, preventing supervision from leaking across control-authority boundaries.

The main contributions of this work are as follows:
\begin{itemize}[leftmargin=*]
    \item \textbf{Hierarchical signal decomposition and dual-head critic.} We show that a single binary episode outcome encodes two separable transition-level signals that require different data and are informative at different training stages.
    We operationalize this decomposition through a viability head $V_v$ and an efficiency head $V_e$, whose one-step advantages are combined via a state-adaptive gate into bounded per-transition weights on the actor loss.
    \item \textbf{Intervention-aware credit assignment.} We restrict outcome labels to policy execution segments according to control authority, preventing credit leakage across intervention boundaries and ensuring clean supervision for both critic heads. 
    \item \textbf{Real-robot validation with open-source release.} On three contact-rich bimanual tasks with deformable objects, HABC improves success rates from 36\%/44\%/12\% to 92\%/88\%/38\% over SFT baselines.
    We will release code and data to facilitate reproducibility.
\end{itemize}

\section{Related Work}
\label{sec:related}
\textbf{Online RL fine-tuning.} HIL-SERL~\citep{luo2025hilserl} and ConRFT~\citep{chen2025conrft} use off-policy RL for robot manipulation, RIPT-VLA~\citep{tan2025riptvla} applies PPO-style updates directly to VLA policies, iRe-VLA~\citep{guo2025irevla} iterates between RL exploration and supervised distillation, VLA-RL~\citep{lu2025vlarl} uses trajectory-level RL with a process reward model for sparse-reward manipulation, and SimpleVLA-RL~\citep{li2025simplevlarl} scales VLA training via RL with a curriculum.
These methods improve task performance through online interaction, but they rely on standard RL machinery and do not focus on how sparse episode outcomes should be converted into transition-level supervision, particularly when policy execution and human intervention are interleaved within episodes.

\textbf{RL for generative action policies.} ReinFlow~\citep{zhang2026reinflow}, FPO~\citep{mcallister2025fpo}, DPPO~\citep{ren2025dppo}, and RFS~\citep{su2026rfs} optimize generative policies via policy-gradient estimation through the generative sampling process, while IDQL~\citep{hansen2023idql} combines implicit Q-learning with diffusion policy extraction.
ARFM~\citep{zhang2026balancing} adaptively balances RL advantage preservation and flow-loss variance for offline post-training of VLA flow models.
Policy-gradient methods require differentiating through the generative process, which can be sample-inefficient for high-dimensional flow-based actors.
HABC takes an alternative approach: it trains critics online with TD learning and converts their outputs into per-transition weights on the supervised flow-matching loss, avoiding policy-gradient estimation entirely while still closing the learning loop through online interaction.

\textbf{Intervention-based imitation learning.} DAgger~\citep{ross2011dagger} and HG-DAgger~\citep{kelly2019hgdagger} aggregate intervention actions as direct supervision.
IWR~\citep{mandlekar2020iwr} increases the weight of intervention data, Sirius~\citep{liu2025sirius} reweights samples using approximated human value judgments, and AIM~\citep{cai2025aim} learns an adaptive criterion for requesting human demonstrations.
RaC~\citep{hu2025rac} scales recovery and correction data through human-in-the-loop rollouts for long-horizon tasks, and MILE~\citep{korkmaz2025mile} models the human intervention decision itself to improve data efficiency.
These approaches treat intervention windows as data to imitate rather than asking how outcomes should be attributed across a mixed-control episode.

\textbf{Advantage-weighted and advantage-conditioned actor updates.}
AWR~\citep{peng2019advantage} and AWAC~\citep{nair2020awac} derive actor weights from critic advantage via $\exp(\hat A/\beta)$; IQL~\citep{kostrikov2021offline}, CQL~\citep{kumar2020cql}, and Decision Transformer~\citep{chen2021decision} provide alternative policy extraction routes from value estimates or returns; AWM~\citep{xue2025awm} analyzes weighted matching losses from a variance-reduction perspective.
In the VLA post-training setting, Recap~\citep{intelligence2025pi} and its scalable system variant SOP~\citep{pan2026sop} convert a reward-derived advantage into a prompt token that conditions the actor, gaining test-time controllability via classifier-free guidance; LWD~\citep{wang2026lwd} extends this to fleet-scale offline-to-online RL with distributed robot experience.
HABC takes the complementary \emph{weighting} route: the actor input is unchanged, and two outcome-label-derived heads produce bounded per-transition loss weights consumed only at training time, decomposing the sparse outcome into viability and efficiency signals.

\section{Method}
\label{sec:method}

\subsection{Problem Setup}
\label{sec:setup}

At each step, the robot observes $s_t = (I_t, q_t, \ell)$, where $I_t$ denotes multi-view images, $q_t$ the proprioception, and $\ell$ a language task prompt,
and executes an action chunk $a_t$ of horizon $H$.
We call a maximal contiguous interval under one controller (current policy or human) a \emph{segment}, a fixed-length training sample drawn from a segment a \emph{window}, and the final policy execution segment after the last intervention a \emph{post-intervention policy execution suffix}.
For notational simplicity, we index each training window by its anchor step $t$.
All methods share a flow-matching VLA actor trained with a weighted flow-matching loss~\citep{lipman2022flow,liu2022rectified,chi2025diffusion,intelligence2025vision}:
\begin{equation}
    \mathcal{L}_{\pi}
    =
    \frac{1}{B}\sum_{i=1}^{B}
    w_i\,
    \bigl\lVert
    v_\theta(s_i,a_i,\sigma)-u_i
    \bigr\rVert_2^2 ,
    \label{eq:weighted_actor_loss}
\end{equation}
where $v_\theta$ is the flow-matching velocity, $u_i$ is the ground-truth flow target, and $w_i$ is a scalar weight derived from the viability value $V_v$ and the efficiency value $V_e$ (\S\ref{sec:critic}).
For readability, Eq.~\eqref{eq:weighted_actor_loss} shows only the scalar transition weight; route-specific validity masks and intervention action-dimension masks are applied in the standard way (details in Appendix~\ref{app:algorithms}).
Online fine-tuning draws from three data sources:
demonstrations $\mathcal{D}_{\mathrm{SFT}}$; autonomous rollouts $\mathcal{D}_{\mathrm{auto}}$ with episode outcome $y\in\{0,1\}$; and human-intervention data $\mathcal{D}_{\mathrm{int}}$.
The core design question is how to set $w_i$ so that viability and efficiency are extracted from sparse outcomes and routed to the correct data, without leaking credit across control-authority boundaries.

\subsection{Dual-Head Critic}
\label{sec:critic}

The scalar weight $w_i$ in Eq.~\eqref{eq:weighted_actor_loss} must encode two distinct improvement signals that are informative at different training stages.
Early in training, when failures are frequent, the key signal is whether an action keeps the task viable.
Later, when success is reliable, the key signal shifts to whether an action advances efficiently.
A single critic trained on episode outcomes conflates these two signals, losing the structure that makes each separately actionable at the right training stage.
We therefore decompose the sparse outcome into two dedicated heads on the shared backbone $\phi(s)$:
\begin{equation}
    z_v(s) = f_v(\phi(s)),
    \quad
    \hat V_e(s) = f_e(\phi(s)),
    \quad
    p_v(s) = \mathrm{sigmoid}(z_v(s)).
    \label{eq:heads}
\end{equation}

$V_v(s)$ estimates the \emph{viability} of state $s$, defined as $p_v(s) = P(\text{success}\mid s)$, the probability of eventual task success under the current policy.
It is trained with binary cross-entropy against the episode outcome $y$ on all labeled policy execution windows.
Because the label itself is binary, $V_v$ can be supervised from both successful and failed episodes, making it informative even when the success rate is low.

$V_e(s)$ estimates the steps to success from state $s$, trained only on successful trajectories where this target is well-defined.
Non-terminal actions receive a one-step cost of $-1$; the terminal success action is assigned target~$0$:
\begin{equation}
    y_e(s_t)
    =
    \begin{cases}
        0, & d_t=1,\\
        -1+\mathrm{sg}[\hat V_e(s_{t+1})], & d_t=0,
    \end{cases}
    \label{eq:epl_target}
\end{equation}
where $d_t=1$ denotes the terminal success step and $\mathrm{sg}[\cdot]$ denotes stop-gradient. $\hat V_e$ is a scalar regression output trained with Huber loss, with target values naturally bounded in $[-T_{\max}, 0]$ where $T_{\max}$ is the episode step limit.
As the policy improves and success becomes frequent, $V_v$ saturates near $1$ for most states; $V_e$ then becomes the more informative ranking signal, distinguishing fast progress from slow progress.

$V_v$ and $V_e$ are trained jointly.
$V_v$ is supervised on all outcome-labeled policy execution windows, including both successful and failed
windows whose outcome is attributable to the current policy.
$V_e$ is supervised only on successful windows, including successful policy-execution windows and successful
intervention windows, because the steps-to-success target is not defined for failures.
Using $\mathcal{D}^{\mathrm{lab}}_{\mathrm{auto}}$ for the former and
$\mathcal{D}_{\mathrm{succ}}=\mathcal{D}_{\mathrm{auto}}^{\mathrm{succ}}\cup
\mathcal{D}_{\mathrm{int}}^{\mathrm{succ}}$ for the latter, the joint critic loss is
\begin{equation}
    \mathcal{L}_{\mathrm{critic}}
    =
    \mathbb{E}_{\mathcal{D}^{\mathrm{lab}}_{\mathrm{auto}}}\!
    \bigl[\mathrm{BCE}(z_v,\,y)\bigr]
    +
    \mathbb{E}_{\mathcal{D}_{\mathrm{succ}}}\!
    \bigl[\mathrm{Huber}(\hat V_e,\,y_e)\bigr].
    \label{eq:critic_loss}
\end{equation}

\subsection{Advantage-Weighted Actor Update}
\label{sec:weighting}
The dual-head critic separates what should be learned from sparse outcomes, but the actor update in
Eq.~\eqref{eq:weighted_actor_loss} requires a scalar weight for each training transition.
We therefore turn each head into a local improvement signal: transitions that increase viability or make efficient
progress are upweighted, while transitions that reduce viability or waste progress are downweighted.
We compute one-step advantages for the two heads and combine them through a state-adaptive gate.

The \emph{viability advantage} measures how much an action improves predicted viability:
\begin{equation}
    A_v = z_v(s_{t+1}) - z_v(s_t) .
    \label{eq:adv_v}
\end{equation}
We compute this difference in logit space to preserve resolution near $p_v\approx 1$.
The \emph{efficiency advantage} measures whether an action shortens the predicted steps to success faster than
the one-step baseline:
\begin{equation}
    A_e = -1 + \hat V_e(s_{t+1}) - \hat V_e(s_t) .
    \label{eq:adv_e}
\end{equation}
Positive $A_e$ indicates the action advances progress beyond expectation; negative $A_e$ indicates slower-than-expected progress.
Both are one-step TD residuals: $A_v$ uses zero per-step reward and measures the change in viability logit over one transition, while $A_e$ incorporates the per-step cost $(-1)$ consistent with the step-count supervision target in Eq.~\eqref{eq:epl_target}.

HABC combines the two advantages through a state-adaptive gate:
\begin{equation}
    g_t = 1+\tanh\!\Bigl(
        \bigl(1-p_v(s_t)\bigr)\,A_v
        +\,p_v(s_t)\,A_e
    \Bigr) .
    \label{eq:viab_weight}
\end{equation}
When $p_v$ is low, the gate emphasizes $A_v$, separating viable states from stuck ones.
When $p_v$ is high, the gate emphasizes $A_e$, ranking actions by efficiency inside already-viable trajectories.
This interpolation happens per state rather than through a global training schedule: within the same batch, a low-viability state is weighted by viability improvement while a high-viability state is weighted by efficiency improvement.

\subsection{Intervention-Aware Credit Assignment}
\label{sec:credit_assignment}

When policy execution and human intervention appear in the same episode, the source of the final outcome is
ambiguous: success may be caused by the policy, by the human correction, or by both.
Naively crediting the outcome to all timesteps leaks supervision across the control-authority boundary.
Concretely, if the episode succeeds after an intervention, naively crediting the outcome to all timesteps would upweight the pre-intervention policy execution segment that led to the near-failure state (reinforcing the mistakes that triggered the intervention); conversely, if the episode fails after an intervention, it would incorrectly penalize the human's corrective actions.

Rather than requiring the critic to infer hidden causes from a binary episode label,
HABC uses the logged control authority as the attribution boundary.
The key observation is that the post-intervention policy execution suffix is the only
segment whose outcome is attributable to the current policy: the pre-intervention
segment led to a state requiring correction, and the intervention segment reflects
human rather than policy decisions.
Corrupting $V_v$ with labels from these segments would cause the viability head to
upweight the very states that triggered failures, undermining the weighting signal.

HABC therefore partitions each episode by controller.
For fully autonomous episodes, the entire trajectory is labeled with $y$.
For episodes with one or more interventions, only the post-intervention policy
execution suffix receives the outcome label.
Intervention windows are never outcome-labeled; instead they serve a dual role:
imitation supervision for the actor and progress targets for $V_e$,
leveraging the human's corrective actions as demonstrations without attributing
the episode outcome to them (see Appendix~\ref{app:credit_assignment} for an illustration).

\subsection{Training Procedure}

We set the pre-normalization weight $\tilde w_i$ according to data source:
$\tilde w_i = g_t$ for successful online rollout samples,
$\tilde w_i = 0$ for failed online rollout samples,
and $\tilde w_i = 1$ for SFT and intervention samples;
when intervention reweighting (IR) is enabled, $\tilde w_i = g_t$ for intervention samples instead.
The normalized weight $w_i$ in Eq.~\eqref{eq:weighted_actor_loss} is then obtained by
unit-mean normalization over valid samples, decoupling the weighting from the effective learning rate:
\begin{equation}
    c=\frac{1}{|\mathcal{D}_{\mathrm{valid}}|}
        \sum_{i\in\mathcal{D}_{\mathrm{valid}}}\tilde w_i,
    \qquad
    w_i=\tilde w_i / \max(c,\varepsilon).
    \label{eq:unit_mean}
\end{equation}
A warmup of $N_{\mathrm{wu}}$ steps keeps $g_t = 1$ until the critic heads are minimally calibrated.
Intervention reweighting is enabled only after an initial HABC phase, once $V_v$ has been trained from labeled policy execution windows collected during that phase.

\section{Experiments}
\label{sec:experiments}

\subsection{Experimental Setup}

\begin{figure*}[h]
    \centering
    \includegraphics[width=\textwidth]{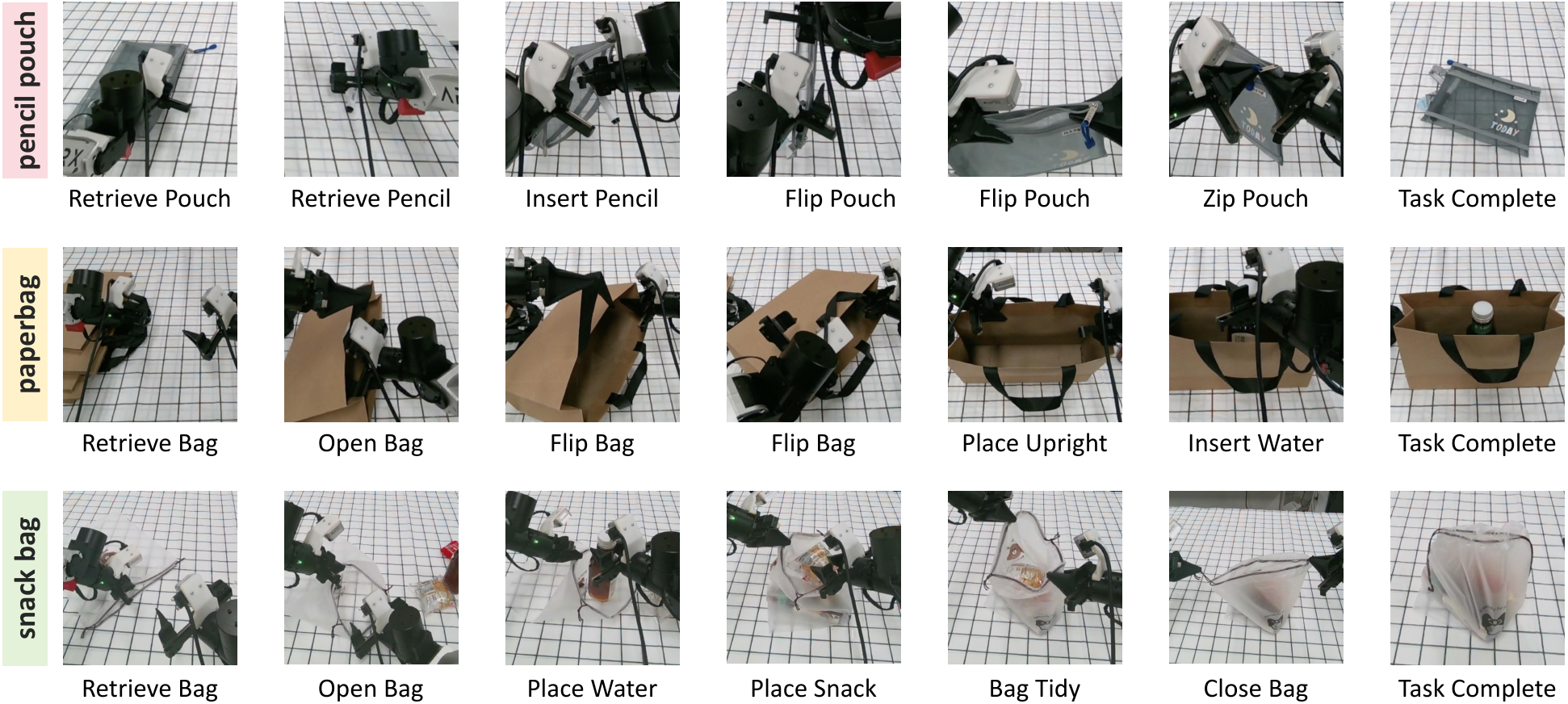}
    \setlength{\abovecaptionskip}{0.5pt} 
    \caption{\textbf{Real-robot bimanual manipulation tasks.}
    We evaluate on three dual-arm tasks involving deformable objects:
    Pencil Pouch,
    Paper Bag and
    Snack Bag.}
    \label{fig:tasks}
\end{figure*}

\textbf{Tasks.}
We evaluate on three real-robot dual-arm manipulation tasks (Figure~\ref{fig:tasks}):
\textbf{Pencil Pouch}, in which the robot inserts a marker and zips a soft pouch closed;
\textbf{Paper Bag}, in which the robot opens a flat-folded bag, stands it upright, and inserts a bottle; and
\textbf{Snack Bag}, in which the robot sequentially places three items into a pouch and pulls the drawstring closed.
All three involve multi-stage bimanual coordination on deformable objects where partial progress and interventions are common.

\textbf{Implementation.}
All experiments are conducted on an ARX X5 bimanual robot.
Observations consist of three RGB camera streams: a top Intel RealSense D455 and two wrist Intel RealSense D405 cameras.
The action space is the robot’s end-effector frame with a chunk size of 50 during training and 25 during inference.
We use $\pi_{0.5}$~\citep{intelligence2025vision} as the base VLA, initialized from its pretrained offline checkpoint, and fine-tune on 8$\times$A800 GPUs.
Full hyperparameter details are in Appendix~\ref{app:hparams}.

\textbf{Baselines.}
We compare five methods: \textbf{SFT} (no online data); \textbf{Imit-DAgger} (50/50 SFT and intervention mix, no rollout data); \textbf{Imit-Recap} (hard-threshold filtering on the critic's TD residual, adapted from~\citep{intelligence2025pi}); \textbf{HABC-V} ($V_v$ only, $V_e$ ablated); and \textbf{HABC} (full method).
Imit-DAgger follows the intervention-imitation recipe of HG-DAgger~\citep{kelly2019hgdagger}; Imit-Recap adopts the hard-threshold filtering mechanism from Recap~\citep{intelligence2025pi} but omits its advantage-conditioned actor prompt, making it a baseline without intervention-aware credit assignment that uses a single critic's TD residual.
The clean ablation of each factor is captured within the HABC variants: HABC-V isolates the contribution of the efficiency head, while the comparison between Imit-Recap and HABC-V highlights the effect of soft dual-head weighting with intervention-aware credit assignment.

\textbf{Metrics and training protocol.}
Each checkpoint is evaluated over 50 trials; we report success rate and mean trajectory length (number of action frames on successful trials only).
For the Step~1 comparison, all online methods are initialized from the same SFT checkpoint and trained with an equal online fine-tuning budget; SFT itself receives no online data and serves as the baseline.
Step~2 starts from the HABC checkpoint and evaluates continued training with and without intervention reweighting.
We additionally report the best checkpoint reached by continuing HABC+IR training for more online rounds, reflecting the full potential of our method given additional interaction data.

\subsection{Main Results}

\begin{figure*}[htbp]
    \centering
    \setlength{\abovecaptionskip}{0.5pt} 
    \includegraphics[width=\textwidth]{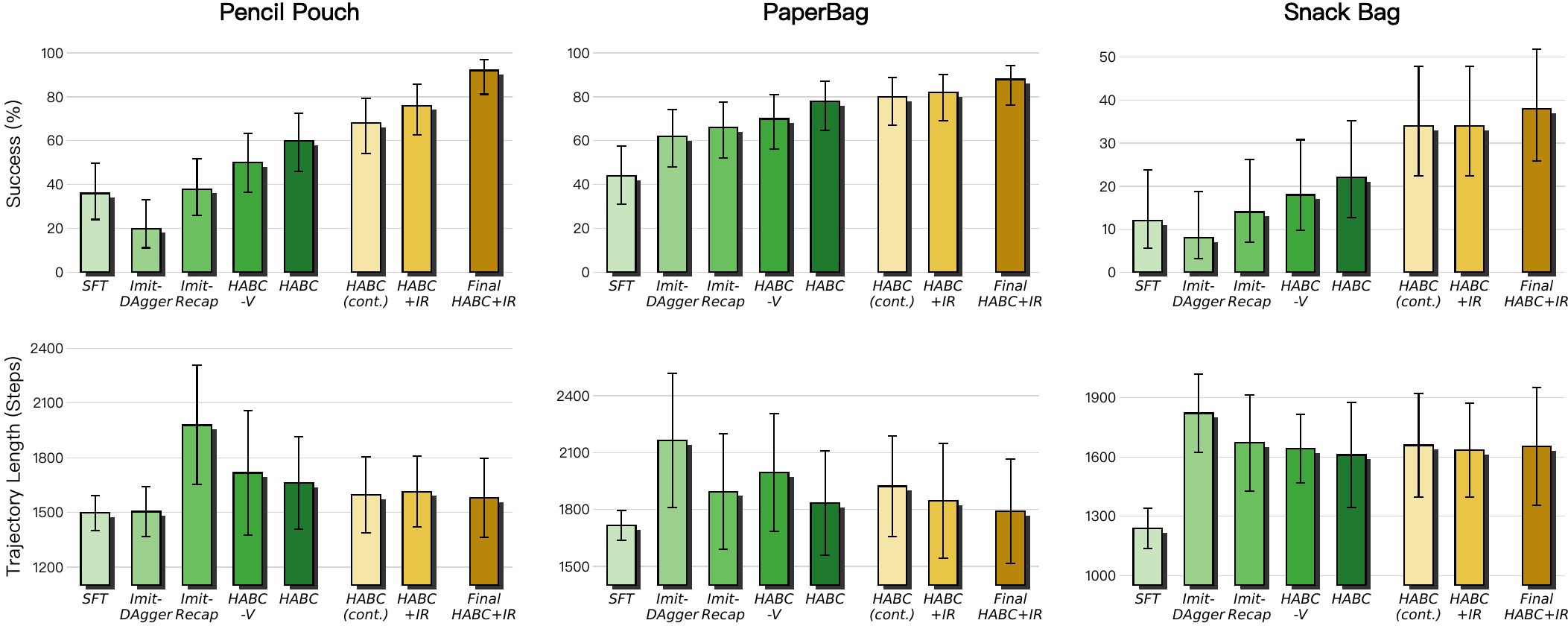}
    \setlength{\abovecaptionskip}{0.5pt} 
    \caption{\textbf{Main results across three tasks.}
    Top: success rate (\%) with Wilson 95\% confidence intervals.
    Bottom: mean trajectory length, measured as number of action frames on successful trials, with standard deviation.
    Methods are grouped into Step~1, initial online fine-tuning with 5 methods; Step~2, continued training $\pm$ intervention reweighting with 2 methods; and the best observed HABC+IR checkpoint.
    Shorter trajectories indicate faster task completion.}
    \vspace{-0.5em}
    \label{fig:main_results}
\end{figure*}

Figure~\ref{fig:main_results} summarizes Step~1, the equal-budget initial online fine-tuning comparison.
HABC achieves the highest success rate on all three tasks (60\%, 78\%, 22\%).
A viability-only variant (HABC-V) already surpasses all non-HABC baselines on every task, suggesting that soft viability weighting extracts more useful supervision from sparse outcomes than hard filtering.
Imit-DAgger underperforms SFT on Pencil Pouch and Snack Bag because the 50/50 mix biases training toward intervention states.

Trajectory length complements success rate by measuring efficiency only among successful trials; it does not conflate efficiency with failure rate.
Upgrading from HABC-V to full HABC consistently reduces trajectory length on every task ($-$55, $-$162, $-$32 frames), suggesting that the efficiency head downweights unproductive motion and favors more direct completions.

Step~2 starts from the HABC checkpoint and compares continued training with and without intervention reweighting.
HABC~(cont.) continues without intervention reweighting, while HABC+IR additionally applies $g_t$ to intervention windows.
Following the training procedure in Section~\ref{sec:method}, intervention reweighting is enabled only after the initial HABC phase.
With multiple additional rounds under HABC+IR, the best checkpoints reach \textbf{92\%} on Pencil Pouch, \textbf{88\%} on Paper Bag, and \textbf{38\%} on Snack Bag, up from SFT baselines of 36\%, 44\%, and 12\%.

\subsection{Critic and Weight Analysis}
\begin{figure*}[htbp]
    \centering
    \begin{minipage}[t]{0.52\textwidth}
        \centering
        \includegraphics[width=\linewidth]{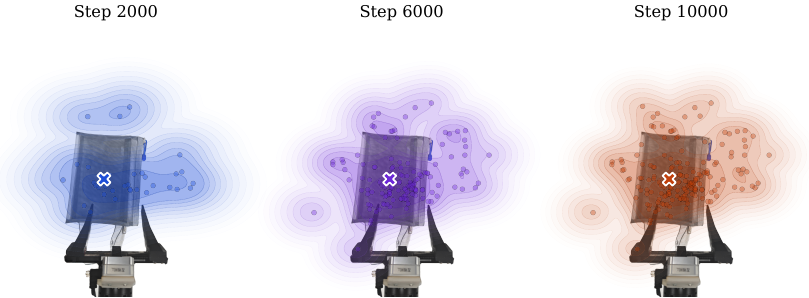}
    \end{minipage}
    \hfill
    \begin{minipage}[t]{0.46\textwidth}
        \centering
        \includegraphics[width=\linewidth]{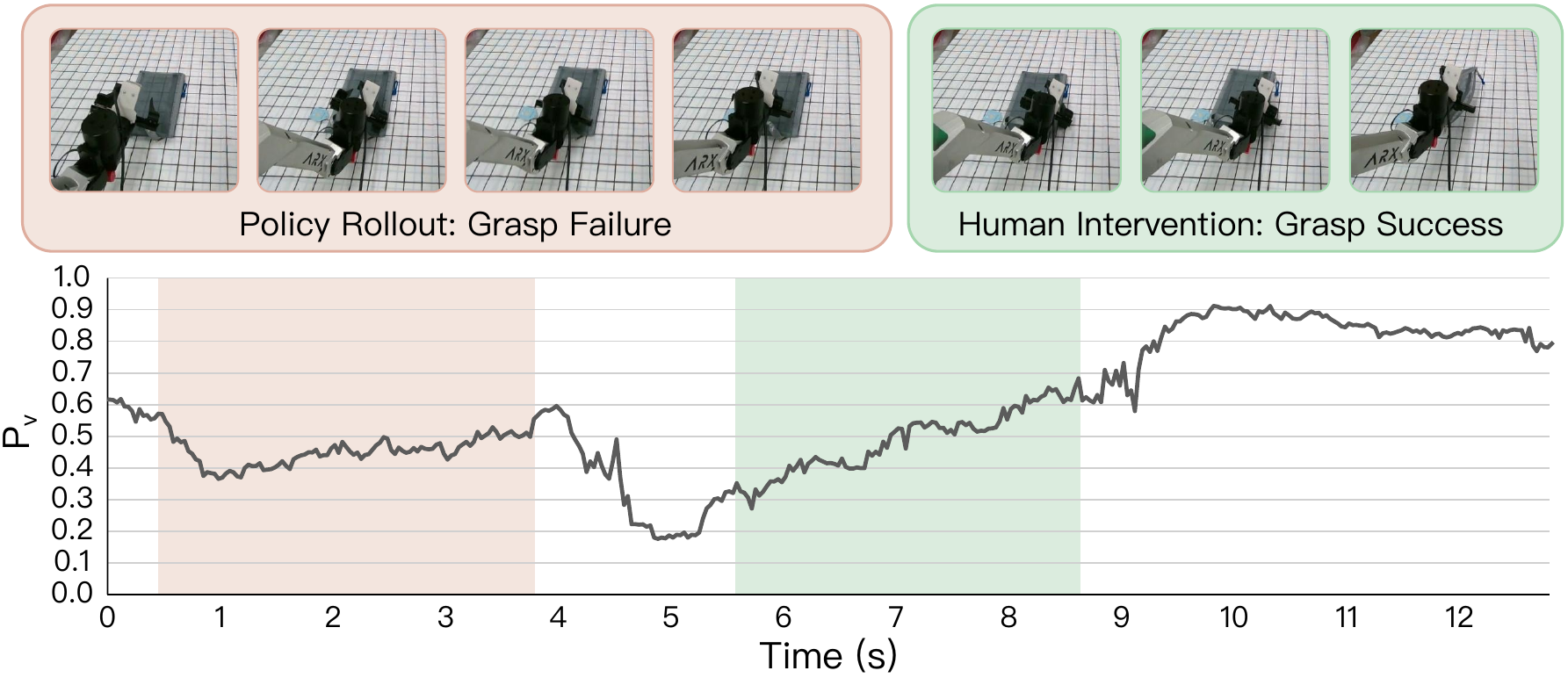}
    \end{minipage}
    \caption{\textbf{Viability head generalization on Pencil Pouch.}
    Left: each point is one episode's initial pouch center position; only positions where $p_v>0.6$ are shown.
    Crosses mark the pouch center.
    As training progresses, the high-viability region expands, indicating that the viability head assigns high viability to an increasingly broad range of observed initial placements.
    Right: along a rollout, $p_v$ drops after an OOD grasp failure and recovers during human intervention.
    This shows that $A_v$ tracks local changes in viability across both autonomous and intervention segments (Section~\ref{sec:weighting}).}
    \vspace{-0.5em}
    \label{fig:succ_diagnostics}
\end{figure*}

\textbf{Value head generalization.}
The viability head $V_v$ provides more than a memorized lookup of training labels (Figure~\ref{fig:succ_diagnostics}).
For Pencil Pouch, we evaluate $p_v$ at each observed initial state, parameterized by the pouch-center position on the workspace; each point represents one episode's initial pouch-center location, and we retain only those where $p_v>0.6$.
As online fine-tuning proceeds, the set of pouch-center positions satisfying $p_v>0.6$ expands progressively (Figure~\ref{fig:succ_diagnostics}, left), indicating that the viability head assigns high viability to an increasingly broad range of observed initial placements.
Trajectory-level traces (Figure~\ref{fig:succ_diagnostics}, right) show the same behavior locally: $p_v$ drops sharply when the policy enters an out-of-distribution state and fails to grasp, then recovers steadily during a human intervention that re-establishes a viable grasp.
The real-time tracking of viability across both autonomous and intervention segments suggests that $A_v$ provides a useful per-transition viability signal for intervention reweighting.

\begin{figure*}[htbp]
    \centering
    \includegraphics[width=\textwidth]{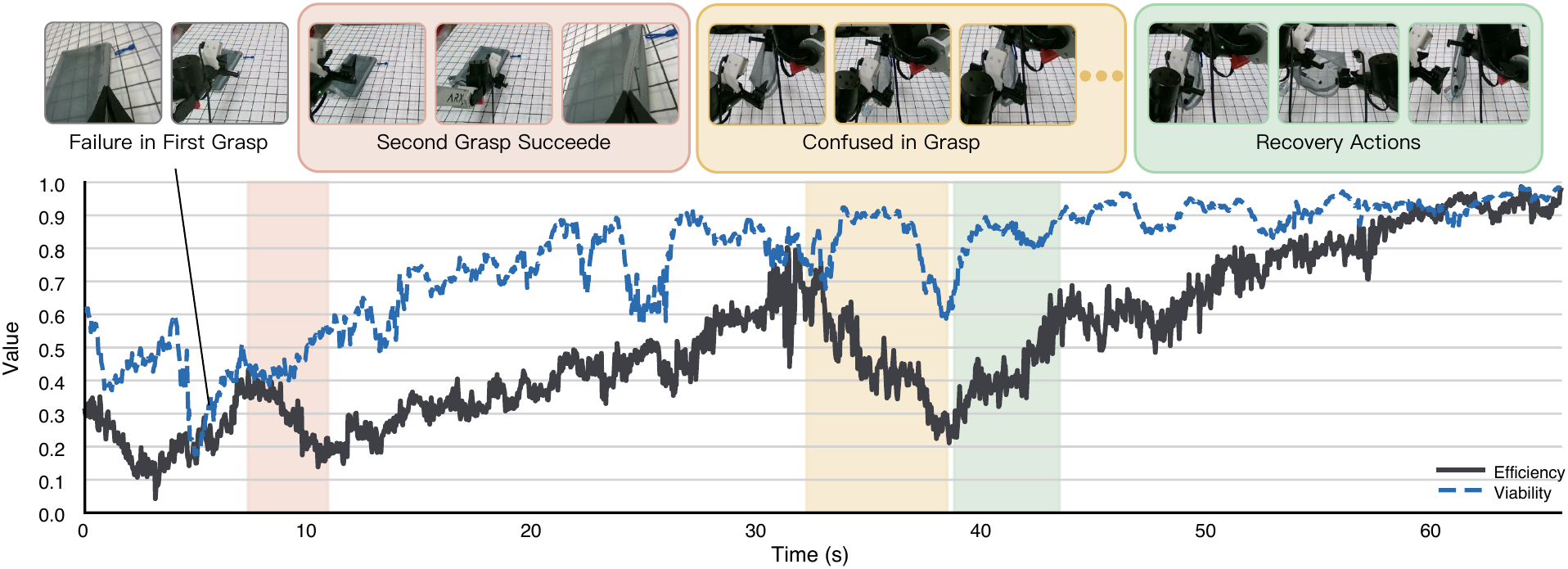}
    \caption{\textbf{Viability and efficiency signals along a Pencil Pouch rollout.}
    Frames at three highlighted segments (pink: successful grasp; yellow: inefficient regrasp; green: recovery) and the corresponding $p_v(s_t)$ (blue dashed) and normalized $\hat V_e(s_t)$ (black solid) over the episode.
    At a successful grasp, $p_v$ rises sharply and the transition is upweighted through $A_v$.
    During inefficient motion, $p_v$ stays high while $\hat V_e$ worsens, so the segment is downweighted through $A_e$.
    During recovery, both signals improve and the transition is upweighted.
    Efficiency rescaled to $[0,1]$ for display.}
    \label{fig:filtering_diagnostics}
\end{figure*}

\textbf{Per-transition weight analysis.}
Figure~\ref{fig:filtering_diagnostics} illustrates how the two heads produce non-uniform weights along a real rollout.
Three highlighted segments demonstrate the division of labor in Eq.~\eqref{eq:viab_weight}.
At a successful second grasp shown in pink, $p_v$ rises sharply and $A_v$ is large, so the gate upweights the transition through the viability term.
During a confused regrasp shown in yellow, $p_v$ stays high and changes little so $A_v$ is uninformative, but $\hat V_e$ signals stalled progress, causing $A_e$ to turn negative and the $p_v A_e$ branch to downweight the inefficient segment.
During recovery actions shown in green, both signals climb together and the gate upweights through both terms.
This complementary behavior, where $V_v$ reacts to discrete viability-changing events while $V_e$ resolves gradations within high-viability regions, helps explain the empirical improvement of HABC over HABC-V in Figure~\ref{fig:main_results}.

\subsection{Recovery Behavior Comparison}
\label{app:recovery_behavior}
\begin{figure*}[htbp]
    \centering
    \includegraphics[width=\textwidth]{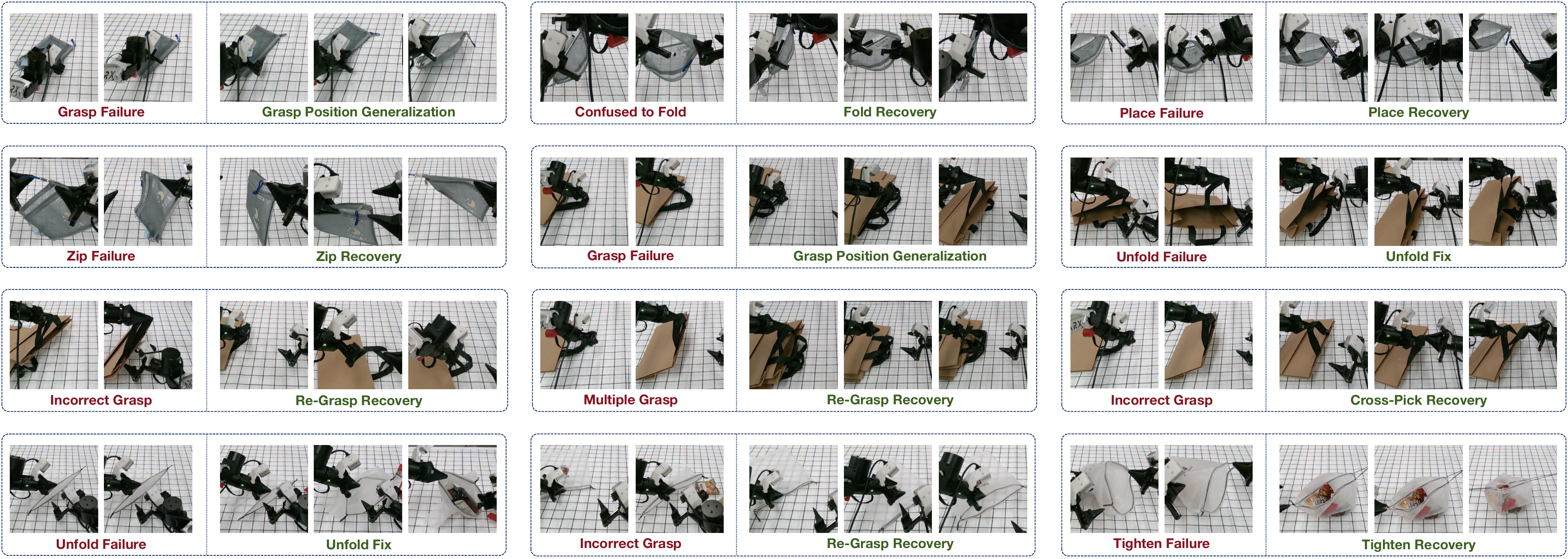}
    \caption{\textbf{Recovery behavior: HABC vs.\ SFT baseline.}
    Each row shows a task where the robot encounters a manipulation failure.
    Top: the SFT baseline fails to recover and the episode terminates unsuccessfully.
    Bottom: the HABC-trained policy detects the error and executes corrective actions, ultimately completing the task.
    The dual-head critic's viability weighting encourages the policy to learn from recovery transitions, enabling autonomous error correction without human intervention.}
    \label{fig:recovery_comparison}
\end{figure*}
Figure~\ref{fig:recovery_comparison} presents qualitative rollout comparisons between the SFT baseline and the HABC-trained policy on three representative failure-recovery scenarios.
In each case, the robot encounters a manipulation error (e.g., a failed grasp, a misaligned insertion, or a dropped object).
The SFT policy, lacking exposure to recovery states during training, either repeats the failed action or enters an unrecoverable loop.
In contrast, the HABC policy---trained with viability-weighted transitions that upweight recovery-oriented actions---detects the failure state and executes corrective motions to re-establish task progress.
These examples complement the quantitative viability-head analysis in Section~\ref{sec:experiments} by showing that the learned weighting translates into observable recovery behavior at deployment.

\section{Conclusion and Limitations}
\label{sec:conclusion}

We presented HABC, an online RL fine-tuning method for VLAs that converts sparse episode outcomes into per-transition behavior-cloning weights via a dual-head critic and intervention-aware credit assignment, leaving the deployed actor unchanged.
The viability head enables learning even when success is rare; the efficiency head reduces trajectory length once success is reliable; and restricting outcome labels to policy execution segments prevents credit leakage across intervention boundaries while preserving human corrections as imitation data.
On three contact-rich bimanual tasks, HABC raises success rates from SFT baselines of 36\%/44\%/12\% to 92\%/88\%/38\%, with qualitative evidence of learned recovery behavior in Appendix~\ref{app:recovery_behavior}.

Intervention-aware credit assignment assumes reliably detected intervention boundaries; noisy labels would corrupt $V_v$ supervision.
$V_e$ trains only on successful trajectories, so its signal is weakest precisely when success is rare.
HABC is currently evaluated on single-task fine-tuning; extending the dual-head design to multi-task or cross-embodiment settings remains an open direction.
In future work, adaptive gating, multi-step advantage estimation, and denser outcome signals for contact-rich recovery are natural next steps to further refine sparse-reward credit assignment.

\bibliography{main}


\clearpage
\appendix

\section{Parent Value Head}
\label{app:parent-value}

The pretrained value head $V_\psi$ follows a distributional formulation~\citep{bellemare2017distributional}.
It predicts logits over value bins and is trained by cross-entropy against a mixed TD/MC target:
\begin{equation}
    y_V(s_t)
    =
    (1-c_t)\bigl(r_t+\gamma\,\mathrm{sg}[V(s_{t+1})]\bigr)
    +
    c_t\,R_t,
    \label{eq:value_target}
\end{equation}
Here $R_t=\sum_{k\geq 0}\gamma^k r_{t+k}$ is the Monte-Carlo return.
The switch variable $c_t\in\{0,1\}$ chooses between the two targets: $c_t{=}1$ applies the MC target at an episode boundary, while $c_t{=}0$ uses the TD target elsewhere.
During online fine-tuning, we continue updating $V_\psi$ on $\mathcal{B}_S\cup\mathcal{B}_O$, starting from the pretrained IQL checkpoint.
This head is retained for compatibility with the Imit-Recap baseline, but it is not used in HABC's actor weighting.
The base VLA model is $\pi_{0.5}$~\citep{intelligence2025vision}.

\section{Compared Update Rules}
\label{app:algorithms}

For completeness, we summarize the update rules used by HABC and the two imitation-style baselines below.

\subsection{HABC}
\label{app:alg-probc}

\begin{algorithm}[htbp]
\caption{One HABC update.}
\begin{algorithmic}[1]
\Require batches $\mathcal{B}_S,\mathcal{B}_I,\mathcal{B}_O$ (sampled
from $\mathcal{D}_{\mathrm{SFT}},\mathcal{D}_{\mathrm{int}},\mathcal{D}_{\mathrm{auto}}$); actor
$\pi_\theta$; parent value head $V_\psi$; critic heads
$f_v,f_e$; warmup $N_{\mathrm{wu}}$;
intervention-reweight flag $\mathrm{IR}$
\Comment{$\mathcal{B}_O^{\mathrm{lab}}=\mathcal{B}_O^{\mathrm{succ}}\cup\mathcal{B}_O^{\mathrm{fail}}$}
\State Update $V_\psi$ on Eq.~\eqref{eq:value_target} over
       $\mathcal{B}_S\cup\mathcal{B}_O$ \Comment{not used in actor weighting}
\State Update $f_v$ using the minibatch estimate of the first term in
       Eq.~\eqref{eq:critic_loss} on $\mathcal{B}_O^{\mathrm{lab}}$
\State Update $f_e$ using the minibatch estimate of the second term in
       Eq.~\eqref{eq:critic_loss} on
       $\mathcal{B}_O^{\mathrm{succ}}\cup\mathcal{B}_I^{\mathrm{succ}}$
\If{step $\geq N_{\mathrm{wu}}$}
    \State Compute $g_t$ from detached head outputs on
           $\mathcal{B}_O^{\mathrm{succ}}$ and, if $\mathrm{IR}$, on
           $\mathcal{B}_I$
\Else
    \State $g_t\leftarrow 1$ for all non-SFT samples
\EndIf
\State Set $\tilde w_i=g_t$ on $\mathcal{B}_O^{\mathrm{succ}}$; set
       $\tilde w_i=0$ on $\mathcal{B}_O^{\mathrm{fail}}$
\State If $\mathrm{IR}$, set $\tilde w_i=g_t$ on $\mathcal{B}_I$; else
       set $\tilde w_i=1$ on $\mathcal{B}_I$
\State Unit-mean-normalize non-SFT weights by Eq.~\eqref{eq:unit_mean}
\State Update $\pi_\theta$ on
       $\mathcal{B}_S\cup\mathcal{B}_I\cup\mathcal{B}_O$ with
       Eq.~\eqref{eq:weighted_actor_loss}
\end{algorithmic}
\end{algorithm}

\subsection{Imit-Recap}
\label{app:filterbc}

Imit-Recap follows the hard-threshold filtering mechanism in Recap~\citep{intelligence2025pi}.
An online transition is included in the actor loss only when the critic's TD residual exceeds a threshold $\epsilon$, selected on a held-out validation set.
This is not a full reproduction of Recap's advantage-conditioned actor.

\begin{algorithm}[htbp]
\caption{One Imit-Recap update.}
\begin{algorithmic}[1]
\Require batches $\mathcal{B}_S,\mathcal{B}_I,\mathcal{B}_O$ (as above); actor
$\pi_\theta$; value head $V_\psi$; threshold $\epsilon$
\State Update $V_\psi$ on Eq.~\eqref{eq:value_target} over
       $\mathcal{B}_S\cup\mathcal{B}_O$
\For{each autonomous rollout sample $i\in\mathcal{B}_O$}
    \State Compute
    $\hat A(s_i,a_i)=r_i+\gamma(1-d_i)V(s_i')-V(s_i)$
    \State Set online weight
    $w_i=\mathds{1}\!\bigl[\hat A(s_i,a_i)\geq\epsilon\bigr]$
\EndFor
\State Set SFT weights to $1$
\State Keep intervention weights unchanged under the intervention mask
\State Update $\pi_\theta$ on
       $\mathcal{B}_S\cup\mathcal{B}_I\cup\mathcal{B}_O$ with
       Eq.~\eqref{eq:weighted_actor_loss}
\end{algorithmic}
\end{algorithm}

\subsection{Imit-DAgger}
\label{app:alg-intbc}

Imit-DAgger follows a simple intervention-imitation recipe.
The actor is trained on a 50/50 mixture of SFT and intervention data, without using rollout transitions or critic-derived reweighting.

\begin{algorithm}[htbp]
\caption{One Imit-DAgger update.}
\begin{algorithmic}[1]
\Require SFT batch $\mathcal{B}_S$; intervention batch $\mathcal{B}_I$;
actor $\pi_\theta$
\State Sample a 50/50 mixed actor batch from $\mathcal{B}_S$ and
       $\mathcal{B}_I$
\State Set SFT weights to $1$
\State Set scalar intervention weights to $1$; use $M_i^{\mathrm{int}}$ as the per-dimension action mask
\State Update $\pi_\theta$ on $\mathcal{B}_S\cup\mathcal{B}_I$ with
       Eq.~\eqref{eq:weighted_actor_loss}
\end{algorithmic}
\end{algorithm}

\section{Intervention-Aware Credit Assignment Illustration}
\label{app:credit_assignment}

\begin{figure*}[htbp]
    \centering
    \includegraphics[width=\textwidth]{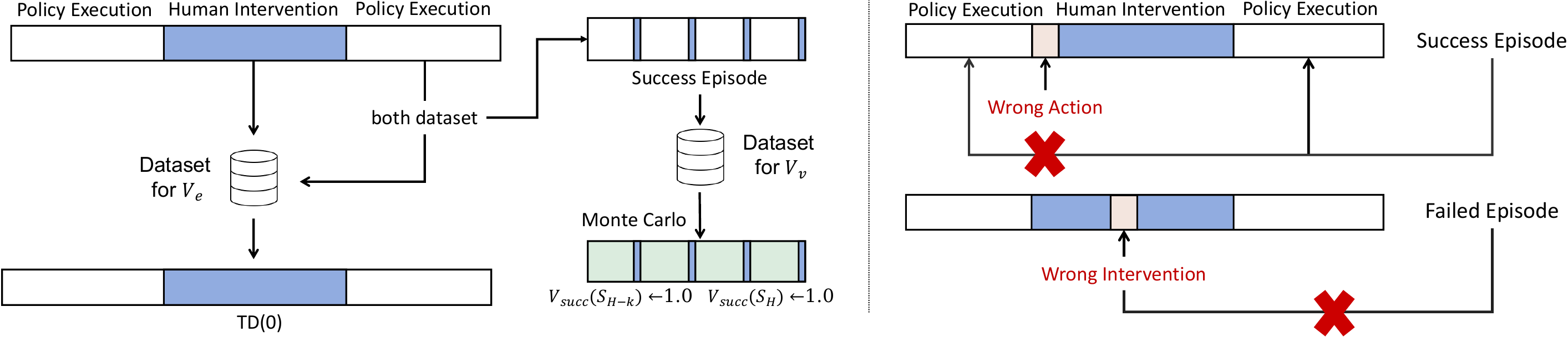}
    \caption{\textbf{Intervention-aware credit assignment and dual-value training.}
    Supervision routing for the two value heads: $V_v$ uses all outcome-labeled policy-execution windows to predict viability, while $V_e$ uses only successful policy or intervention windows to predict progress/efficiency.
    Their advantages become transition-level actor weights.
    In online episodes, the source of the final success/failure label is uncertain, so naive episode-level supervision can assign credit to the wrong controller.}
    \label{fig:credit_assignment}
\end{figure*}

\section{Windowing Details}
\label{app:windowing}

For each episode that contains an intervention, we split the trajectory into policy execution segments and intervention segments.
If an episode contains an intervention, only the post-intervention policy execution suffix receives the episode outcome label.
This suffix is the part executed by the current policy from a corrected state onward.
Earlier policy execution segments are kept in the replay buffer but do not receive outcome labels.
For fully autonomous episodes, the full trajectory receives the episode outcome label.
An intervention window requires at least 10 human-controlled steps within a 50-step window.
This intervention-aware split is used for both critic supervision and actor weighting.

\section{Data Collection Protocol}
\label{app:data_protocol}

Each task starts from 200 SFT demonstration episodes.
Online fine-tuning then proceeds in rounds.
Each round collects 100 autonomous rollout episodes and trains for 6k gradient steps.
Among failed rollouts, approximately half receive human intervention, where the operator takes over and completes the task.
The remaining failed rollouts are recorded as unassisted failures.
Each round therefore adds 100 autonomous rollout episodes to the replay buffer, consisting of successes, episodes with intervention, and pure failures.

For Pencil Pouch, the initial-phase best checkpoint in Figure~\ref{fig:main_results} is selected after 3 online rounds, corresponding to 300 rollout episodes on top of 200 SFT demonstrations.
Continued training with HABC+IR then runs for additional rounds.
The final HABC+IR checkpoint reaches 92\% success.
For Paper Bag, the same schedule applies: 3 initial rounds followed by continued HABC+IR rounds, with a final best checkpoint of 88\%.
For Snack Bag, we again use 3 initial online rounds followed by continued HABC+IR rounds, with a final best checkpoint of 38\%.

\section{Hyperparameters}
\label{app:hparams}

Key constants are $C=100$ (episode failure penalty; applied as reward $r=-C$ on failed-episode transitions for the parent value head TD update (Appendix~\ref{app:parent-value}) and the Imit-Recap advantage computation—HABC's actor weighting does not use this reward directly), $N_{\mathrm{wu}}=500$ (warmup steps), Huber $\delta=1.0$, $\gamma=0.99$, batch size $B=256$, and a stale-rollout cutoff of 10000 model indices.
All HABC runs use pure TD supervision for $V_e$ and keep action-expert training disabled.
Online fine-tuning is initialized from the pretrained $\pi_{0.5}$ IQL checkpoint.

\section{Weight Statistics}
\label{app:weight_stats}

To verify that the HABC weighting rule produces meaningful variation rather than near-uniform weights, we report empirical weight statistics from the three tasks.

After warmup, the mean pre-normalization weight $g_t$ on successful autonomous rollout transitions is approximately $0.76$, $0.68$, and $0.86$ for Pencil Pouch, Paper Bag, and Snack Bag respectively.
These values indicate that the weighting rule is non-trivial and varies across tasks.

For comparison, Imit-Recap passes approximately 22, 27, and 19 out of every 100 sampled autonomous rollout transitions for Pencil Pouch, Paper Bag, and Snack Bag respectively.
The remaining transitions are discarded by the hard threshold.
This filtering behavior is more aggressive than HABC's soft weighting.

When intervention reweighting is enabled, the mean $g_t$ on intervention windows is approximately $1.0$, $1.1$, and $0.95$ for Pencil Pouch, Paper Bag, and Snack Bag respectively.
On average, the critic therefore assigns near-uniform weight to intervention windows.
Individual intervention transitions still receive non-uniform weights, but the distribution is more concentrated than for autonomous rollout data, consistent with the expectation that interventions are generally productive.

In the main text, mean trajectory length is reported only over successful evaluation trials.
We use this metric as a direct readout of trajectory efficiency: once a policy can solve the task, fewer frames indicate less redundant motion and less recovery before completion.
The consistent reductions from HABC-V to HABC in Figure~\ref{fig:main_results} therefore support the intended role of $V_e$.


\end{document}